\documentclass[conference]{IEEEtran}
\IEEEoverridecommandlockouts
\usepackage{cite}
\usepackage{amsmath,amssymb,amsfonts}
\usepackage{algorithmic}
\usepackage{graphicx}
\usepackage{cleveref}
\usepackage{textcomp}
\usepackage{xcolor}
\def\BibTeX{{\rm B\kern-.05em{\sc i\kern-.025em b}\kern-.08em
    T\kern-.1667em\lower.7ex\hbox{E}\kern-.125emX}}
\begin{document}

\title{Beyond Sliders: Mastering the Art of Diffusion-based Image Manipulation}

\author{%
\textbf{Yufei Tang}$^{1}$ \quad \textbf{Daiheng Gao}$^{2}$ \quad \textbf{Pingyu Wu}$^{2,3}$ \quad \textbf{Wenbo Zhou}$^{2,3}$ \quad \textbf{Bang Zhang}$^{4}$ \quad \textbf{Weiming Zhang}$^{2,3}$\\
$^1$FUYAO University Of Science And Technology\quad $^2$University of Science and Technology of China \\ \quad $^3$Anhui Province Key Laboratory of Digital Security \quad $^4$Alibaba TongYi Lab \\\
$^1$786650170@qq.com\quad $^2$samuel.gao023@gmail.com \\
}

\maketitle

\begin{abstract}
In the realm of image generation, the quest for realism and customization has never been more pressing. While existing methods like concept sliders have made strides, they often falter when it comes to non-AIGC images, particularly images captured in real-world settings. To bridge this gap, we introduce \textbf{Beyond Sliders}, an innovative framework that integrates GANs and diffusion models to facilitate sophisticated image manipulation across diverse image categories. Improved upon concept sliders, our method refines the image through fine-grained guidance—both textual and visual—in an adversarial manner, leading to a marked enhancement in image quality and realism. Extensive experimental validation confirms the robustness and versatility of \textbf{Beyond Sliders} across a spectrum of applications.
\end{abstract}


\section{Introduction}
In the field of image generation, the ability to control and edit images is crucial. One pioneering approach to image editing has been through Generative Adversarial Networks (GANs), as introduced in the seminal work by Goodfellow \textit{et al.} \cite{goodfellow2014generative}. Since then, GANs have been extensively utilized for a multitude of image generation and editing tasks.

However, GANs have encountered certain limitations, notably in the precision of image control. This challenge has spurred researchers and practitioners to devise innovative methods to enhance control over image creation. Among these advancements, the StyleGAN series \cite{Alpher04, Alpher05, karras2021alias} stands out as particularly renowned.


With the advent of diffusion models, generative modeling has opened new frontiers\cite{Alpher06}. Particularly, diffusion models using latent space have demonstrated exceptional capabilities in generating high-quality images. Unlike GANs, diffusion models gradually transform random noise into coherent images through a series of denoising steps\cite{Alpher08}. This iterative process not only produces realistic images but also showcases intricate details and textures\cite{Alpher09}.

As for image editing, current diffusion-based SOTA: Concept Sliders\cite{Alpher12,han2024parameter} enable fine control over specific features, such as age, gender, thus becoming a powerful tool for personalized image generation and editing\cite{Alpher05,Alpher12}, advancing the boundaries of image editing\cite{Alpher07}. However, despite these advancements, unresolved issues remain when testing images of natural scenes. Natural scene images, also known as in-the-wild images, captured in uncontrolled environments, typically contain various lighting conditions, backgrounds, and subject poses. Editing these images presents unique challenges due to their inherent diversity and complexity, and poor generalization performance hinders Concept Sliders from landing.

\begin{figure}[t]
\centering
\includegraphics[width=0.5\textwidth]{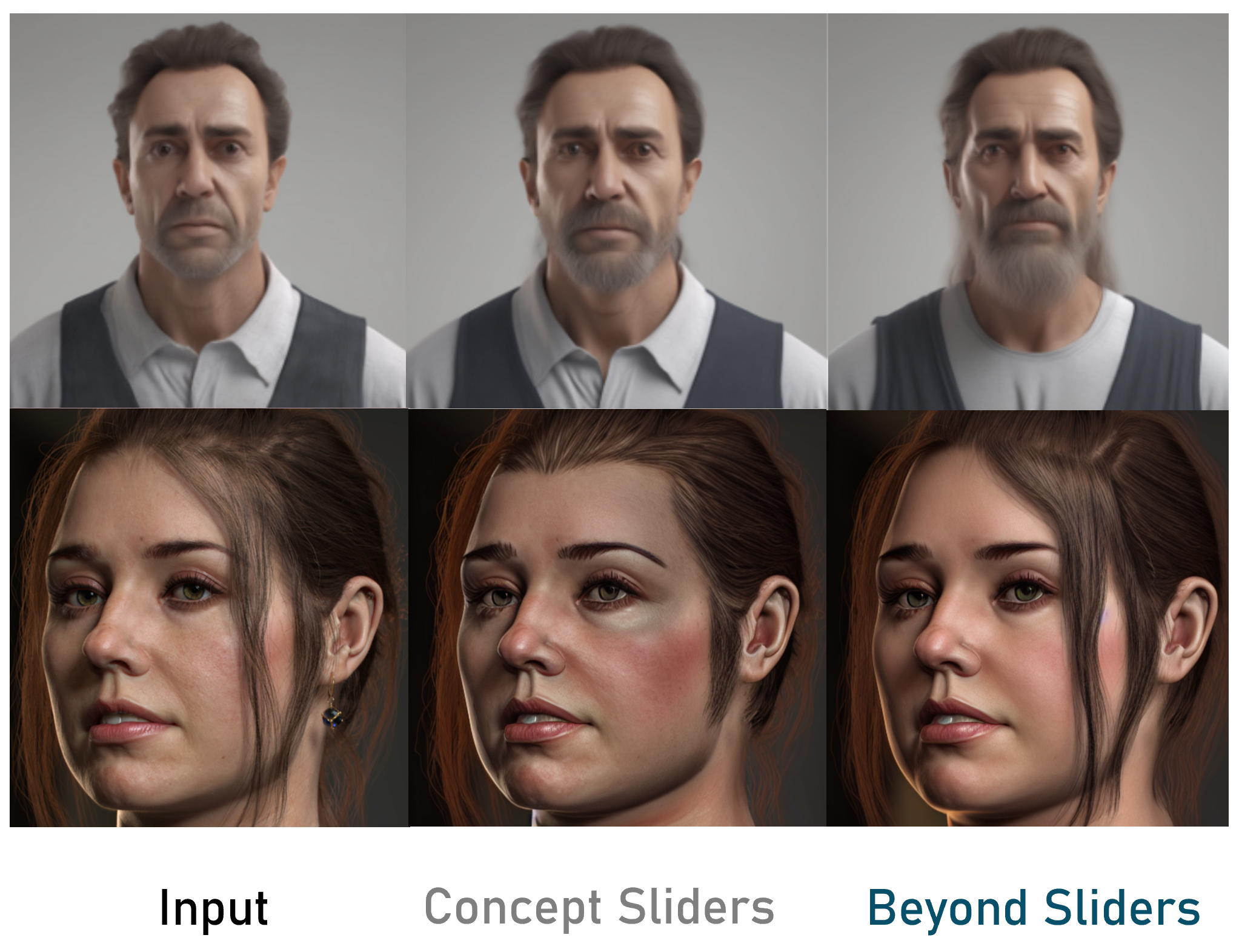}
\caption{Beyond Sliders evolved upon its precedant, Concept Sliders,  in terms of identity-consistency and high-fidelity. }
\label{fig:teaser}
\vspace{-0.2in}
\end{figure}

As shown in \cref{fig:teaser}, we introduce \textbf{Beyond Sliders}, which harnesses the power of GANs and diffusion models to achieve elaborative image manipulation on images of various types. Specifically, by utilizing Low-Rank Adaptation (LoRA)\cite{hu2021lora} within the stable diffusion framework\cite{rombach2022high}, our method optimizes the generated images by innovatively combining a discriminator network and perceptual loss\cite{johnson2016perceptual}. This approach ensures excellent image quality and realism, even when applied to complex and diverse natural scene images.

Overall, \textbf{Beyond Sliders} addresses the limitations of previous techniques, such as Concept Sliders, which struggle with the diversity and complexity of in-the-wild images. The adoption of conceptual triplet loss is also a key element in our work, which ensures the low-rank directions in diffusion latent space for targeted concept control. 


\begin{figure*}[htbp]
\centering
\includegraphics[width=1.0\textwidth]{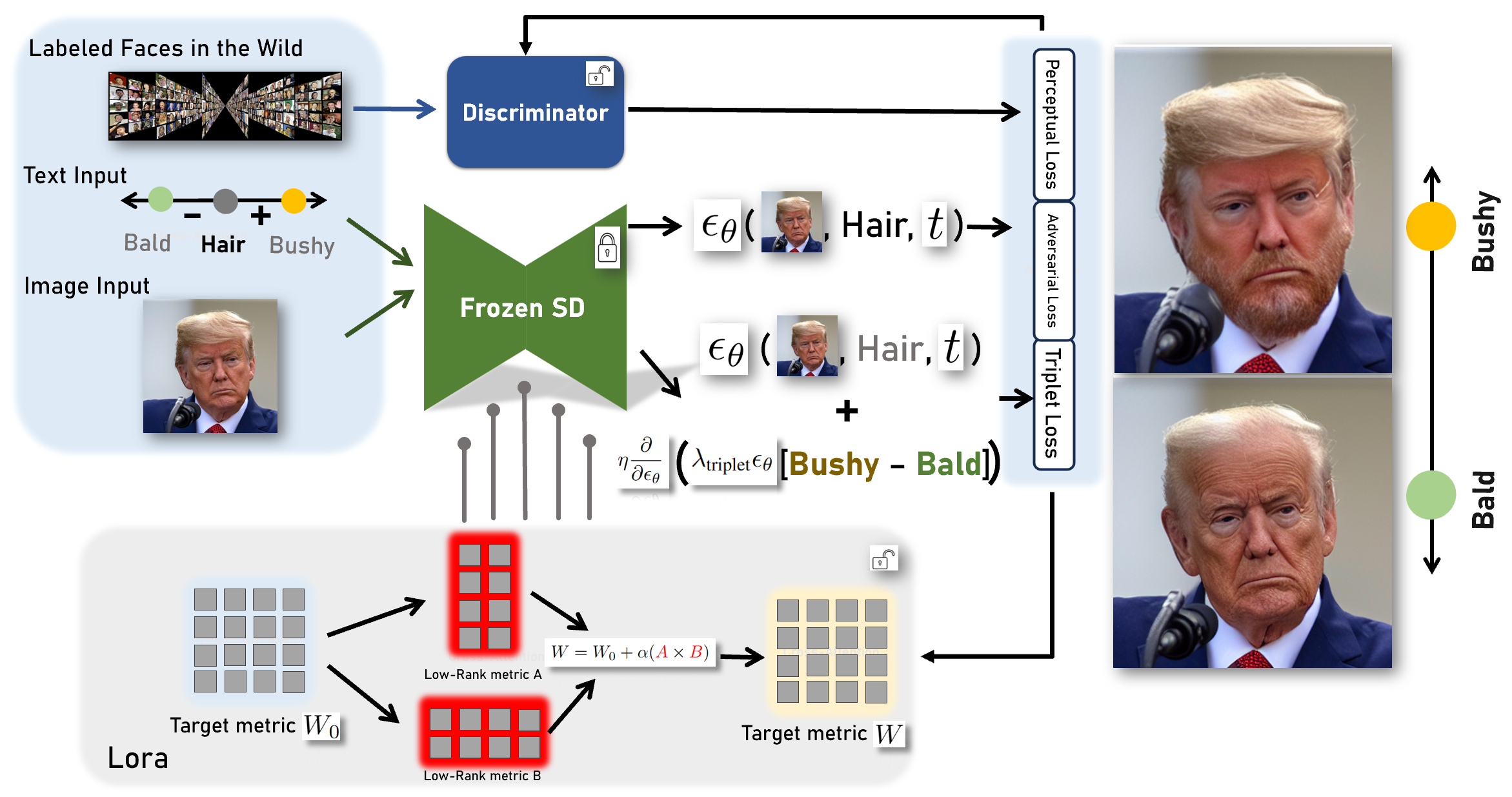}
\caption{\textbf{Beyond Slider's Workflow}, utilizing a frozen SD and LoRA adapters to process image inputs. The model is optimized through perceptual loss, adversarial loss, and triplet loss to improve the generated outputs. Text inputs like "Bald" and "Bushy" guide the model to generate images with specific attributes. Additionally, weight updates are performed using LoRA, allowing for precise control over both the quality and attributes of the generated images.}
\label{fig:method}
\vspace{-0.2in}
\end{figure*}

\section{Related Works}

Conditional image editing is gaining popularity, focusing on creating or adjusting images based on certain conditions. While GANs have been considered, we prioritize diffusion-based methods for their reliable and high-quality results.

Reversion\cite{huang2023reversion} innovatively extracts abstract concepts from a few images, simplifying the process of concept-based editing with diffusion models. Concept Sliders\cite{Alpher12}, which employs LoRA\cite{hu2021lora} for precise control in diffusion models, represents a cutting-edge approach and seminal work in conditional image editing topic. By fine-tuning low-rank adapters, it allows for precise control over image attributes with minimal interference. 

Notable developments based on Concept Sliders include TexSliders\cite{Alpher21}, which adds detailed texture editing, and SOEDiff\cite{Alpher23}, which enhances small object inpainting with advanced techniques. SCoFT\cite{Alpher22} ensures diverse image generation through self-contrastive fine-tuning. These methods reflect the ongoing progress towards more intuitive and manageable image editing tools.

By the way, our method fundamentally differs from existing latent space\cite{parihar2022everything} manipulation techniques in several ways:

\textsc{I. Integration of Triplet Loss}: Unlike GANSpace\cite{harkonen2020ganspace} or linear SVM\cite{Vladimir1995SVM} approaches, our method incorporates conceptual triplet loss to explicitly align latent directions with target attributes, improving semantic control and precision.

\textsc{II. Focus on Generalization}: While prior methods like StyleFlow\cite{StyleFlow2021} often rely on specific attribute classifiers, Beyond Sliders leverages low-rank adaptation (LoRA) and adversarial training to generalize across diverse image types, including in-the-wild images.

\textsc{III. Diffusion Models}: Existing methods are typically limited to GAN frameworks. Our work integrates diffusion models, leveraging their iterative refinement process for enhanced realism and flexibility.

\textsc{IV. Perceptual Loss and Adversarial Training}: Our method leverages perceptual loss and adversarial training to enhance image quality and semantic alignment. Perceptual loss captures realism through feature-level differences, while adversarial training teaches the model to learn complex distributions, overcoming issues like mode collapse and unnatural artifacts. Together, they enable Beyond Sliders to balance semantic fidelity with realism, facilitating complex tasks like attribute disentangling and editing real-world images.


\section{Method}

As shown in \cref{fig:method}, we utilized Stable Diffusion XL\cite{Alpher12} as our base model, enhanced by LoRA\cite{hu2021lora}, perceptual loss\cite{johnson2016perceptual} and GAN\cite{Alpher02} to improve the realism and consistency of generated images. The model consists of a generator (SDXL + LoRA) and a discriminator, trained together in an adversarial manner to ensure high-quality image generation.

\subsection{Preliminaries}

\textbf{Diffusion Models} Diffusion models are a class of generative models that operate by reversing a diffusion process to generate data. Initially, the forward diffusion process gradually adds noise to the data, transitioning it from an organized state $x_0$ to a complete Gaussian noise state $x_T$. At any timestep $t$, the noised image is modeled as:

\begin{equation}
x_t \leftarrow \sqrt{1 - \beta_t} x_0 + \sqrt{\beta_t} \epsilon
\end{equation}

where $\epsilon$ is randomly sampled Gaussian noise with zero mean and unit variance. Diffusion models aim to reverse this diffusion process by sampling a random Gaussian noise $X_T$ and gradually denoising the image to generate an image $x_0$. The objective of the diffusion model is simplified to predicting the true noise $\epsilon$  by minimizing the following loss function\cite{Alpher06,calvin2022understand}:

\begin{equation}
\nabla_{\theta} \| \epsilon - \epsilon_{\theta}(x_t, c, t) \|^2
\end{equation}

where $\epsilon_{\theta}(x_t, c, t)$is the noise predicted by the diffusion model at timestep $t$ conditioned on $c$. In our work, we use SDXL\cite{rombach2022high, podell2023sdxl}, which improve efficiency by operating in a lower dimensional latent space $z$ of a pre-trained variational autoencoder (VAE)\cite{kingma2013auto}. Finally, the latent vector $z_0$ is decoded through the VAE decoder to get the final image $x_0$.

\textbf{LoRA Adaptors} To further enhance control over image generation, we use Low-Rank Adaptation (LoRA) to fine-tune the model\cite{hu2021lora}. LoRA decomposes the weight update $\Delta W$ of large-scale pre-trained models into low-rank matrices $A$ and $B$, significantly reducing the number of trainable parameters. The weight update is defined as:

\begin{equation}
\Delta W = A \times B
\end{equation}

where $B \in \mathbb{R}^{d \times r}$, $A \in \mathbb{R}^{r \times k}$, and $r \ll \min(d, k)$. By freezing the original weights $W_0$ and only optimizing the smaller matrices $A$ and $B$, LoRA achieves massive reductions in trainable parameters. During inference, the final weight is represented as:

\begin{equation}
W = W_0 + \alpha \Delta W
\end{equation}

Increasing $\alpha$ during inference strengthens the edit without retraining the model.

\subsection{Triplet Loss with $c_+$, $c_-$, and $c_t$}

In our method, we adjust the model's generation control by leveraging a triplet loss\cite{schroff2015facenet,hermans2017defense,hoffer2015deep} involving $c_+$ (positive conception, e.g. \texttt{bushy, old}), $c_-$ (negative conception, e.g. \texttt{bald, young}), and $c_t$ (neutral conception, e.g. \texttt{hair, age}), which enhances or suppresses specific attributes in the generated image. Given a target concept $c_t$ and model $\theta$, we aim to find $\theta^*$ that modifies the likelihood of attributes $c_+$ and $c_-$ in image $X$. This increases the likelihood of $c_+$ and decreases the likelihood of $c_-$ conditioned on conception $c_t$.

We achieve this by the following update:

\begin{equation}
P_{\theta^*}(X | c_t) \leftarrow P_{\theta}(X | c_t) \left( \frac{P_{\theta}(c_+ | X)}{P_{\theta}(c_- | X)} \right)^{\eta}
\end{equation}

Where $P_{\theta}(X | c_t)$ represents the distribution generated by the original model when conditioned on $c_t$. Expanding $P(c_+ | X) = \frac{P(X | c_+) P(c_+)}{P(X)}$, we derive the log-probability gradient:

\begin{equation}
\begin{split}
\nabla \log P_{\theta^*}(X | c_t) + \lambda ( \nabla \log (P_{\theta}(X | c_+) - P_{\theta}(X | c_-))
\end{split}
\end{equation}

To further strengthen control over the generation process, we introduce \textit{perceptual loss} into the triplet loss framework. This helps to preserve the similarity between the generated images and reference images at the feature level, providing additional supervision signals during training.

The perceptual loss function is defined as:

\begin{equation}
L_{\text{perp}} = \frac{1}{N} \sum_{i=1}^{N} \| \phi(G(X_i)) - \phi(X_{\text{real}, i}) \|^2
\end{equation}

Where $\phi(\cdot)$ represents the feature extraction function from a pre-trained network (such as VGG~\cite{simonyan2014VGG}), $G(X_i)$ is the generated image, and $X_{\text{real}, i}$ is the corresponding real image.

The combined loss function is expressed as:

\begin{equation}
L_{\text{total}} = \lambda_{\text{triplet}} \cdot L_{\text{triplet}} + \lambda_{\text{perp}} \cdot L_{\text{perp}}
\end{equation}

We introduce a dynamic weight adjustment strategy. Early in training, the weight of the perceptual loss is large, while the weight of the triplet loss gradually increases as training progresses to ensure semantic consistency in the generated images. The weight adjustment strategy is as follows:

\begin{equation}
\lambda_{\text{perp}}(t) = \frac{1}{1 + e^{k(t-t_0)}}
\end{equation}
\begin{equation}
\lambda_{\text{triplet}}(t) = 1 - \lambda_{\text{perp}}(t)
\end{equation}

Where $t$ represents the training step, $t_0$ denotes the switching phase, and $k$ controls the speed of the transition. This dynamic adjustment strategy enables the model to generate high-quality images in the early stages of training, while ensuring semantic consistency in the later stages.

The optimization of the denoising prediction is performed at each timestep $t$, by adjusting the noise prediction $\epsilon_{\theta}(X, c_t, t)$ to minimize the total loss function $\mathcal{L}_{\text{total}}$. Thus, the optimization of denoising prediction is essentially carried out under the framework of the total loss function:


\begin{equation}
\epsilon_{\theta^*}(X, c_t, t) \leftarrow \epsilon_{\theta}(X, c_t, t) + \eta \frac{\partial \mathcal{L}_{\text{total}}}{\partial \epsilon_{\theta}}
\end{equation}

By doing so, the model can gradually optimize the generated images to ensure that they possess good visual quality and meet the expected semantic goals. 


\subsection{Adversarial Training}

To further enhance the realism and quality of generated images, we introduce adversarial training: a generator and a discriminator\cite{lecun1989bCNN,Alpher02}. The discriminator is a binary classifier used to distinguish between real images and generated images, with real images sourced from the LFW dataset\cite{huang2008labeled}, and the generator is the combination of SDXL + LoRA. The generator is trained to produce images that "fool" the discriminator, and the generated images are pitted against real images in the discriminator, resulting in adversarial loss and perceptual loss\cite{johnson2016perceptual,simonyan2014VGG}. These losses are fed back to optimize the weights of both the generator and the discriminator simultaneously, which provides additional supervisory signals and enhances the visual quality of the generated images, helping the model avoid mode collapse.

Adversarial training helps address the issue where models may lose detail or realism when focusing solely on semantic accuracy (e.g., measured by CLIP). With adversarial training, we can maintain the visual quality and semantic relevance of the generated images.

\section{Experiments}

In this section, we conducted multiple experiments to assess the performance and efficacy of our approach. These include evaluations across text-based and real-world scenarios, studies on rectifying flawed image outputs, and composition trials.

\begin{figure}[t]
\centering
\includegraphics[width=0.5\textwidth]{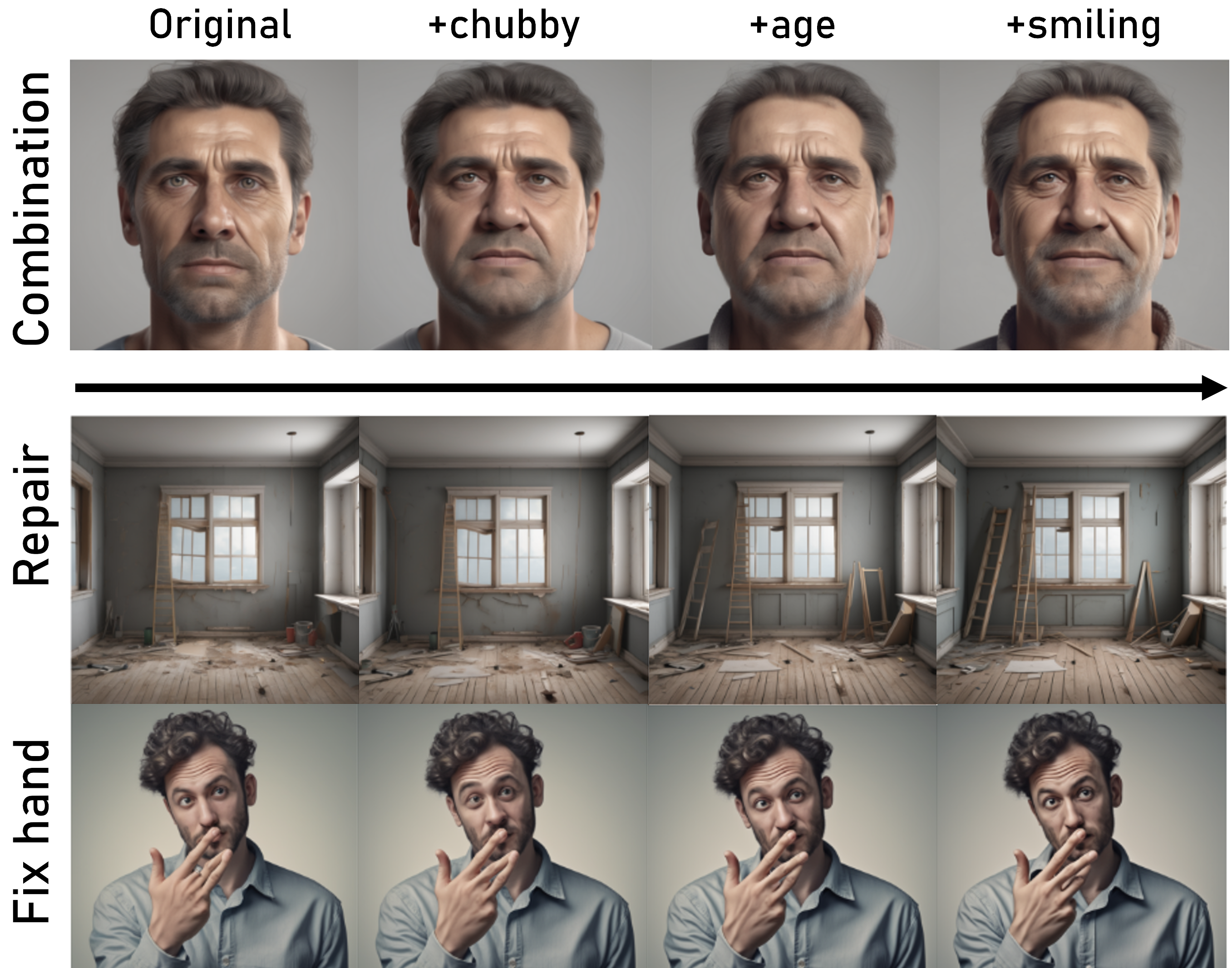}
\caption{We show the composition capability through progressively compose multiple sliders from left to right (\textbf{Top row}), and demonstrate beyond sliders is capable to steer highly abstract concepts: hand and indoor layout to be physically plausible (\textbf{Bottom rows}).}
\label{fig:exp2}
\vspace{-0.2in}
\end{figure}

\subsection{Text Guided \& In-the-wild Testing}

\begin{figure}[t]
\centering
\includegraphics[width=0.5\textwidth]{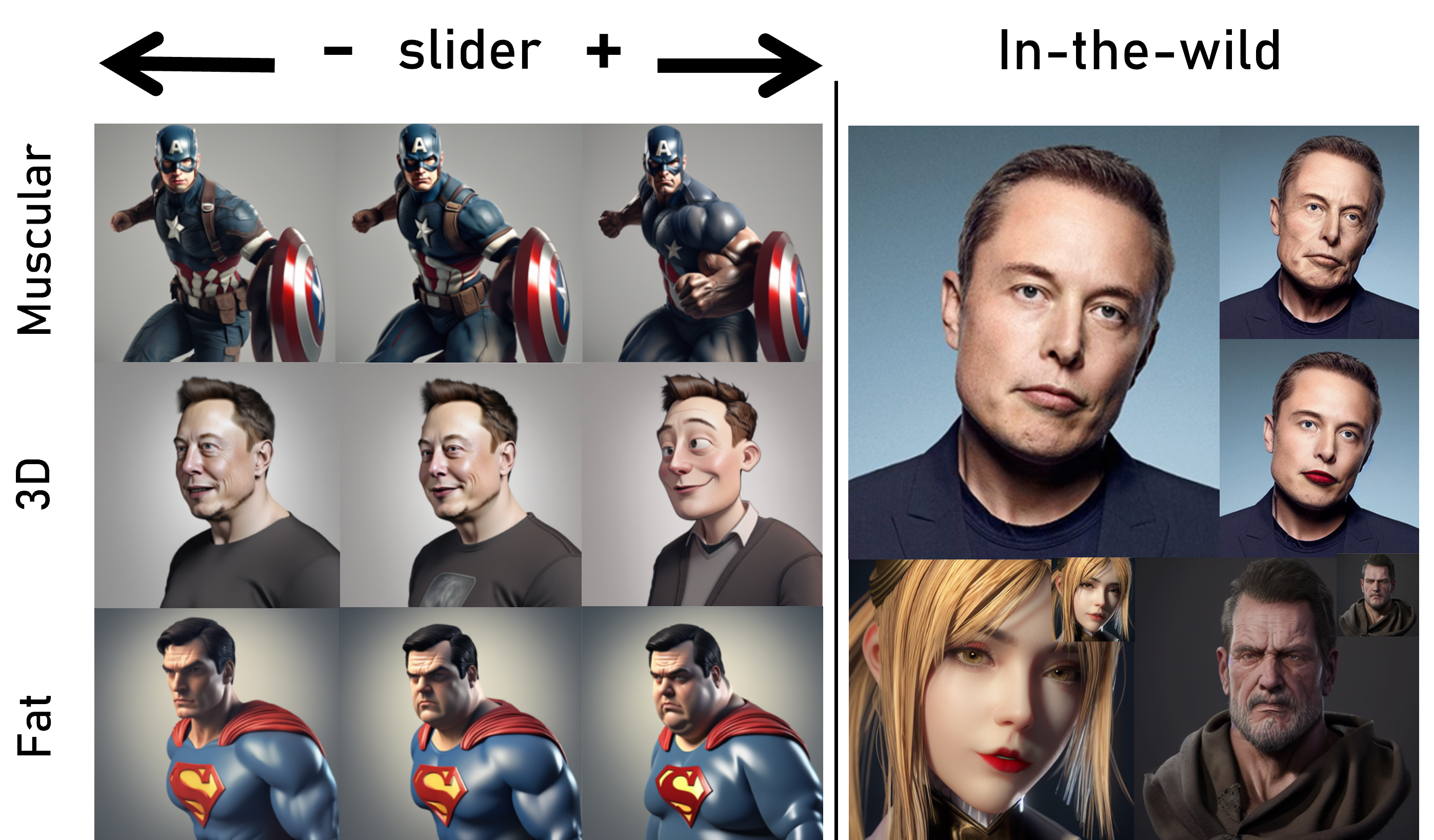}
\caption{Beyond sliders enable meticulous adjustment of specific image attributes during the generation process, while demonstrate strong generalization on in-the-wild data, including 3D game character, celebrity.}
\label{fig:exp1}
\vspace{-0.18in}
\end{figure}

In ~\cref{fig:exp1}, we noticed that our method is adept at: 1) altering identities across both low-level (Fat) and high-level (3D) abstractions, and 2) robustly generalizing to diverse real-world exemplar data, encompassing everything from actual celebrity photos to 3D characters sourced from video games. 

As we observed, the generated images were more natural in appearance, without any noticeable flaws. The modifications made to the attributes were both clear and striking. This demonstrates that Beyond Sliders is not only proficient in maintaining the fine details and naturalness of images but also shows an enhanced ability to manage sophisticated attribute alterations effectively.

Additionally, we evaluated the generated images from both models using CLIP and LPIPS scores. The results show that the images generated by our model achieved higher CLIP scores and lower LPIPS scores, indicating that our model produces more realistic image edits and is closer to the original images.~\cite{Alpher28,Alpher29}

\subsection{Slider Composability \& Erroneous Correction}
To further illustrate the disentanglement capabilities of our sliders, we selected three specific sliders: "chubby," "age," and "smiling," for our composable experiments. The rationale behind choosing these sliders is that the visual changes they induce are both intuitive and easily discernible.

\begin{figure}[h]
\centering
\includegraphics[width=0.5\textwidth]{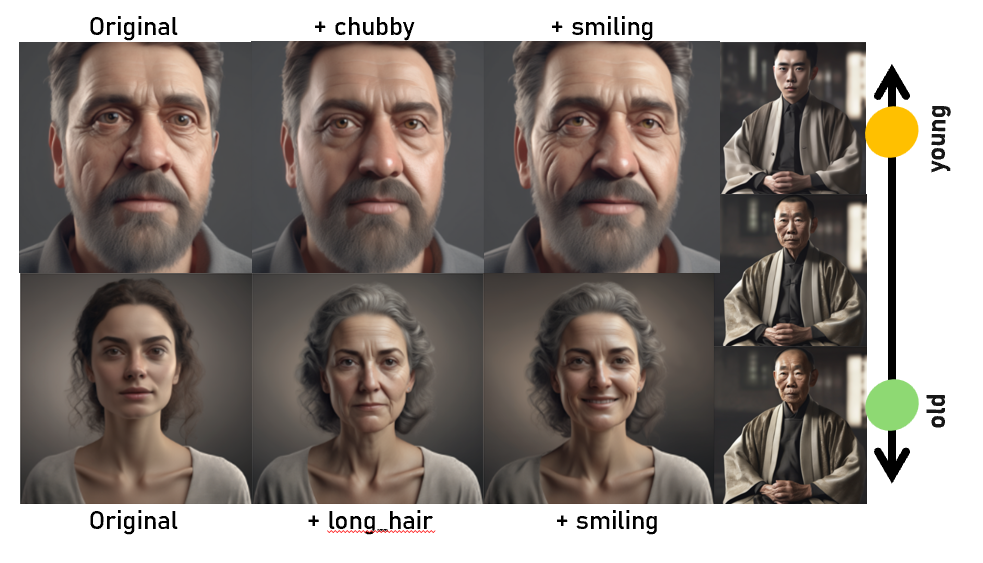}
\caption{We showcases supplementary results from the slider composability tests, underscoring that the model's efficacy is far from coincidental.}
\label{fig:supp1}
\vspace{-0.2in}
\end{figure}

Throughout the experiment, we manipulated these sliders at varying degrees to incrementally alter the image generation outcomes. The process begins with the creation of a base image derived from the initial textual prompt. Subsequently, the "chubby" slider was applied to the base image. Following this, the "age" slider was introduced on top of the "chubby" effect to produce an intermediate image. Lastly, the "smiling" slider was superimposed onto the image, culminating in the final output. As depicted in the top row of \cref{fig:exp2} and left part of \cref{fig:supp1}, the final image, when juxtaposed with the base image, clearly exhibits the cumulative effects of the three sliders applied.

In the bottom rows of~\cref{fig:exp2}, by leveraging sliders, our model is capable of rectifying unnatural or distorted elements within the image: bad hands and anomalistic room.  

As the slider scales are increased, the model progressively generates hands that are more accurate, realistic, and proportionally correct.


\begin{table}[htbp]
\vspace{-0.5em}
\caption{quantitative comparison with state-of-the-art method}
\begin{center}
\label{tab:qqq}
\begin{tabular}{ccccc}
\hline
 Category & Weight & Model & CLIP $\uparrow$ & LPIPS $\downarrow$ \\
\hline
Text Guided & long\_hair  & \textit{Concept Sliders}  &  29.03 & 0.041   \\
            & long\_hair  & \textit{Our Method}  &  \textbf{29.06} & \textbf{0.026}   \\
            & muscular  & \textit{Concept Sliders}  & 28.58 & 0.091  \\
            & muscular  & \textit{Our Method}  & \textbf{28.61} & \textbf{0.063}  \\
            & age  & \textit{Concept Sliders}  &  \textbf{25.90} & 0.038    \\
            & age  & \textit{Our Method}  &  25.84 & \textbf{0.027}    \\
\hline
In the Wild Texting & age & \textit{Concept Sliders} & 21.92 & 0.051 \\
                    & age & \textit{Our Method} & \textbf{22.15} & \textbf{0.047} \\ 
                    & chubby  & \textit{Concept Sliders} & \textbf{26.80} & 0.078 \\
                    & chubby  & \textit{Our Method}  & 26.77 & \textbf{0.037} \\
\hline
\end{tabular}
\end{center}
\end{table}

\subsection{Quantitative Results}



We evaluated the generated images from both models using CLIP\cite{Alpher28} and LPIPS\cite{Alpher29}. As shown in the results~\cref{tab:qqq}, for text-guided image generation tasks across various weights, Beyond Sliders and Concept Sliders show notable differences:

For slider "muscular" and "age", Beyond Sliders shows a clear advantage in LPIPS, particularly in the age task, where Beyond Sliders achieves an LPIPS loss of 0.027 compared to 0.038 for Concept Sliders.

Specifically, with slider "long hair", Beyond Sliders achieved a CLIP score of 29.06 and an LPIPS loss of 0.026, compared to Concept Sliders' 29.03 CLIP score and 0.041 LPIPS. This indicates that Beyond Sliders not only produces images more aligned with the text description but also maintains better visual consistency. 


In the In-the-Wild image editing task~\cref{tab:qqq}, we tested the editing capabilities of Beyond Sliders and Concept Sliders on specific attributes such as age and body type. The results show that, for the age task, Beyond Sliders achieved a CLIP score of 22.15 and an LPIPS loss of 0.047, which is better than Concept Sliders' 21.92 CLIP score and 0.051 LPIPS. In the meantime, for the chubby task, Beyond Sliders demonstrated a significant advantage in LPIPS, scoring 0.037, which is considerably lower than Concept Sliders' 0.078. This suggests that Beyond Sliders offers superior performance in preserving visual consistency during image edits.


\begin{table}[htbp]
\caption{Ablation Study on loss configs}
\begin{center}
\label{tab:ablation_phase}
\begin{tabular}{ccc}
\hline
Setup & CLIP $\uparrow$ & LPIPS $\downarrow$ \\
\hline
Full set & \textbf{34.89} & \textbf{0.557} \\
\texttt{w/o adv}  & 32.19 & 0.586 \\
\texttt{w/o perp} & 34.43 & 0.644 \\
\texttt{w/o adv \& perp} & 33.17 & 0.629 \\
\hline
\end{tabular}
\vspace{-0.5em}
\end{center}
\end{table}

\subsection{Ablation study}
Here, we ablate the two key parts of our method: (1) \textbf{perceptual loss} \texttt{perp} and (2) \textbf{adversarial training} \texttt{adv}. 

\begin{figure}[t]
\centering
\includegraphics[width=0.46\textwidth]{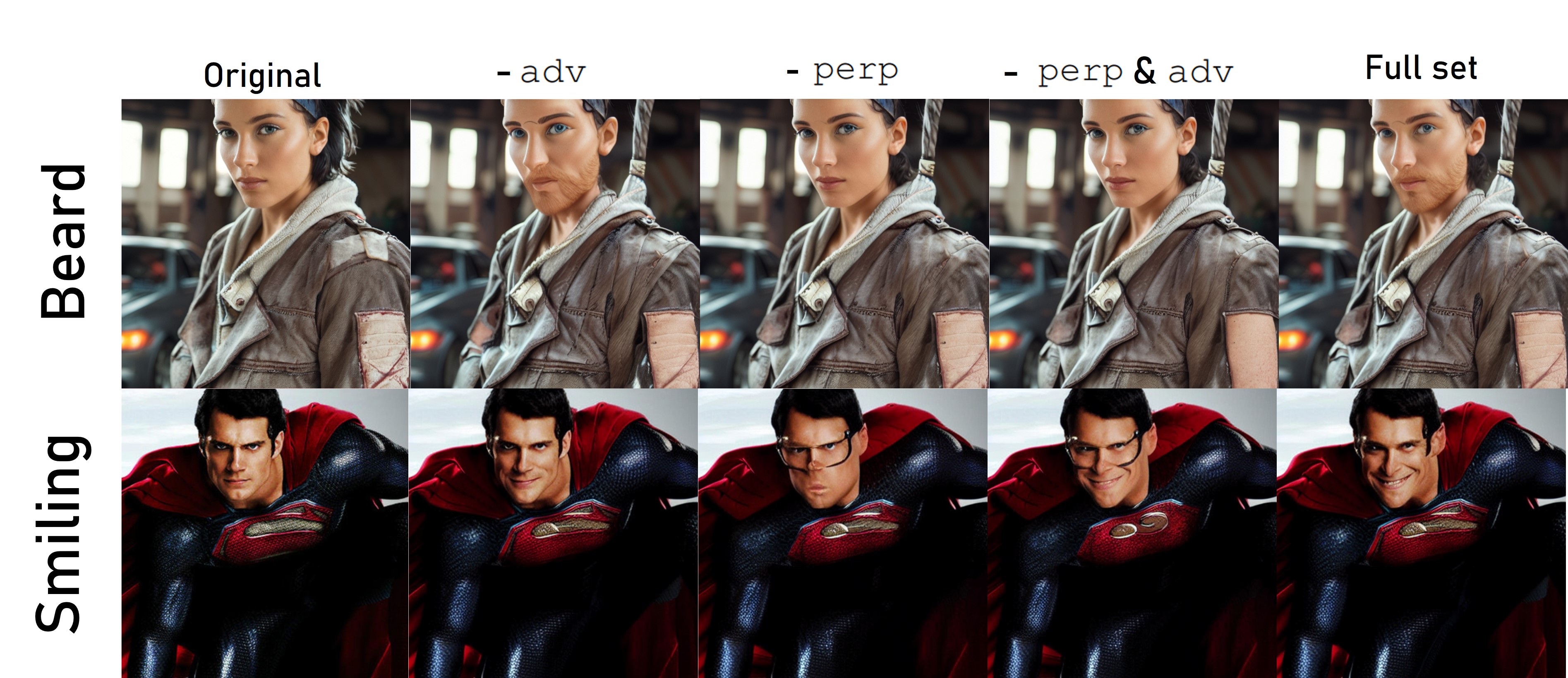}
\caption{Ablation study on loss configs.}
\label{fig:exp4}
\vspace{-0.2in}
\end{figure}

As shown in \cref{fig:exp4} and \cref{tab:ablation_phase}, perceptual loss helps maintain fine-grained details and ensures that the generated images closely resemble the original reference in terms of visual features. On the other hand, adversarial training contributes to enhancing the overall realism and naturalness of the generated images by encouraging the model to produce outputs that can "fool" a discriminator network. This combination allows for the generation of images that are not only semantically accurate but also visually coherent and lifelike. 

The results showed that removing different modules led to varying degrees of decline in image quality. When perceptual loss was removed, the generated images exhibited distortion and were far from resembling real images. When adversarial training was removed, the generated images lost the supervision of adversarial training, leading to reduced contrast, and the overall generation was inferior to the full model. The worst results occurred when both modules were removed, leading to higher distortion and images that were significantly different from the original ones.

We also measured the CLIP and LPIPS scores at each stage of the ablation experiments. The results showed that, under the full model setup, the CLIP score reached 34.9, and the LPIPS score was 0.557. This indicated that the full model achieved a high CLIP score, meaning the generated images had strong semantic alignment with the text descriptions, while the low LPIPS score demonstrated a high degree of visual similarity between the generated images and the reference images. After removing adversarial training, the CLIP score dropped significantly to 32.2, and the LPIPS score increased to 0.586. When perceptual loss was removed, the CLIP score dropped to 34.4, and the LPIPS score rose to 0.644. When both the discriminator and perceptual loss modules were removed, the CLIP score decreased to 33.2, and the LPIPS score increased to 0.629, indicating the importance of both components in generating high-quality images.



\section{Limitation}

In this study, we primarily focus on the field of conditional image editing, especially methods based on diffusion models. However, there are still the following limitations:

1.\textbf{Lack of experiments on the latest Flow-Matching models}: We have not yet conducted experiments on the latest Flow-Matching models (e.g., Flux~\footnote{https://blackforestlabs.ai/}, Stable Diffusion 3~\cite{sd3}). These models exhibit significant advantages in generation quality and efficiency. Future research should consider incorporating them into experiments to evaluate their performance in conditional image editing tasks.

2.\textbf{Limited generalization ability}: Although our method has achieved success in facial editing tasks, its performance on other image categories has not been fully validated. This limits the model's widespread applicability. Future work should focus on improving the model's generalization ability across different image categories to meet diverse editing needs.

To address these limitations, future research should explore the following directions:

1.\textbf{Integrating the latest Flow-Matching models}: Incorporating models such as Flux and Stable Diffusion 3.5 into experiments to assess their effectiveness in conditional image editing.

2.\textbf{Improving the model's generalization ability}: Developing editing methods suitable for various image categories to ensure high editing performance in different scenarios.

By addressing these issues, we aim to further advance the field of conditional image editing, providing more versatile and efficient editing tools.

\section{Conclusion}

In this paper, we present Beyond Sliders, a versatile technique for exerting clear and understandable control over diffusion models. It distinguishes itself from Concept Sliders by adeptly capturing the nuances of semantic directions within the latent space, facilitating an intuitive and granular adjustment of image features across both AIGC-generated and real-world imagery. A pivotal advancement is the incorporation of adversarial training, which significantly elevates image quality—a fact substantiated by our extensive experimental results.

\bibliographystyle{IEEEbib}

\end{document}